# Implementing Systemic Thinking for Automatic Schema Matching: An Agent-Based Modeling Approach

Hicham Assoudi, Hakim Lounis

Département d'Informatique
Université du Québec à Montréal
Succursale Centre-ville, H3C 3P8, Montréal, Canada
Email: assoudi.hicham@courrier.uqam.ca    lounis.hakim@uqam.ca

*Abstract*— Several approaches are proposed to deal with the problem of the Automatic Schema Matching (ASM). The challenges and difficulties caused by the complexity and uncertainty characterizing both the process and the outcome of Schema Matching motivated us to investigate how bio-inspired emerging paradigm can help with understanding, managing, and ultimately overcoming those challenges. In this paper, we explain how we approached Automatic Schema Matching as a systemic and Complex Adaptive System (CAS) and how we modeled it using the approach of Agent-Based Modeling and Simulation (ABMS). This effort gives birth to a tool (prototype) for schema matching called Reflex-SMAS. A set of experiments demonstrates the viability of our approach on two main aspects: (i) effectiveness (increasing the quality of the found matchings) and (ii) efficiency (reducing the effort required for this efficiency). Our approach represents a significant paradigm-shift, in the field of Automatic Schema Matching.

## I. INTRODUCTION

Schema Matching is an important task for many applications, such as data integration, data warehousing and e-commerce. Schema matching process aims at finding a pairing of elements (or groups of elements) from the source schema and elements of the target schema such that pairs are likely to be semantically related [2] [3].

Schema matching existing approaches rely largely on human interactions, either for the matching results validation, during the post-matching phase, or for the matching process optimization, during the pre-matching phase. Although this human involvement in the automatic matching process could be considered as acceptable in a lot of matching scenarios, nevertheless it should be kept to a minimum, or even avoided, when dealing with high dynamic environments (i.e., semantic Web, Web services composition, agents communication, etc.) [1]. Thus, the existing approaches are not suited for all the matching contexts due to their intrinsic limitations. We can summarize those limitations as follows:

- Lack of autonomy to the extent that the user involvement is still needed for the results validation and analysis, but also for matching process configuration and optimization (tuning) to improve the matching result quality and then reduce uncertainty.
- Lack of adaptation in sense that the optimization task of the matching tool must be repeated and adapted manually, for every new matching scenario.

We were motivated to investigate other prospects not yet applied on Schema Matching. We try to answer the following general question: "How can we, with the help of a generic approach, better manage complexity and uncertainty inherent to the automatic matching process in general, and in the context of dynamic environments (minimal involvement of the human expert)?"

More specifically, we asked the following questions:

(i) How can we model the complexity of the matching process to help reduce uncertainty?

(ii) How can we provide the matching process of autonomy and adaptation properties with the aim to make the matching process able to adapt to each matching scenario (self-optimize)?

(iii) What would be the theoretical orientation that may be adequate to respond to the above questions?

In our work, we have investigated the use of the theory of CAS emanating from systemic thinking, to seek, far from the beaten path, innovative responses to the challenges faced by classical approaches for automatic schema matching, (e.g., complexity, uncertainty). The central idea of our work is to consider the process of matching as a CAS and to model it using the approach of ABMS. The aim being the exploitation of the intrinsic properties of the agent-based models, such as emergence, stochasticity, and self-organization, to help provide answers to better manage complexity and uncertainty of Schema Matching.

Thus, we propose a conceptual model for a multi-agent simulation for schema matching called Schema Matching as Multi-Agents Simulation (SMAS). The implementation of this conceptual model has given birth to a prototype for schema matching (Reflex-SMAS).

Our prototype Reflex-SMAS was submitted to a set of experiments, to demonstrate the viability of our approach with respect to two main aspects: (i) effectiveness (increasing the quality of the found matchings), and (ii) efficiency (reducing the effort required for this efficiency). The results came to demonstrate the viability of our approach, both in terms of effectiveness or that of efficiency.

The empirical evaluation results, as we are going to show in Section IV of this paper, were very satisfactory for both effectiveness (correct matching results found) and efficiency (no optimization needed to get good result from our tool).

The current paper is organized as follows: Section 2 discusses schema matching through a state of the art that identifies the important factors affecting the schema matching process. Section 3 presents the chosen paradigm to address the problem. Section 4 shows the results obtained by our approach, and how we can compare them to those obtained in other works. Finally, the last section concludes and summarizes this work.

## II. CURRENT APPROACHES OF SCHEMA MATCHING

Many algorithms and approaches were proposed to deal with the problem of schema matching and mapping [1] [4]–[15]. Although the existing schema matching tools comprise a significant step towards fulfilling the vision of automated schema matching, it has become obvious that the user must accept a degree of imperfection in this process. A prime reason for this is the enormous ambiguity and heterogeneity of schema element names (descriptions). Thus, it could be unrealistic to expect a matching process to identify the correct matchings for any possible element in a schema [16] [17].

A comprehensive literature review, of the existing matching tools and approaches, allowed us to identify the most important factors affecting, in our opinion, the schema matching process. Moreover, some causal relationships, between those different factors, participating to the schema matching difficulties and challenges, were identified. As shown in Figure 1, the factors influencing the Schema Matching are:

- Heterogeneity: in general, the task of matching involves semantics (understanding the context) to have complete certainty about the quality of the result. The main challenge in all cases of automatic matching is to decide the right match. This is a very difficult task mainly because of the heterogeneity of the data.
- Uncertainty: the cause for this uncertainty lies mainly in the ambiguity and heterogeneity, both syntactic, and semantic, which often characterize the Schema Elements to match.
- Optimization: the uncertainty about the matching results implies the optimization of the process to improve the matching quality, and the testing of different combinations (e.g., different Similarity Measures, Aggregate Functions, and Matching Selection Strategies). Each step of the matching process involves choosing between multiple strategies, which leads to a combinatorial explosion (complexity).
- Complexity: the matching process optimization generates complexity because of the search space (combinatorial explosion). In addition, changing matching scenarios exacerbates this complexity to the extent that the result of the optimization often becomes obsolete with changing scenarios.

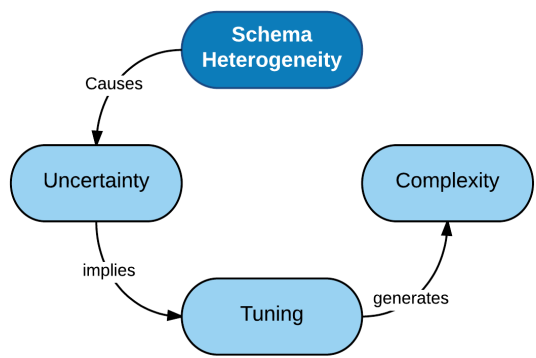

Figure 1. Schema Matching impacting factors causality diagram

One of the commonalities between all existing approaches is the thinking behind these approaches, namely reductionism (as opposed to holism). The reductionist thinking is a very common and efficient thinking approach. It is at the basis of the almost totality of previous schema matching approaches, and then, on their characteristics that are, in our view, the root causes preventing the automatic matching schemes to cope fully with the challenges and difficulties.

Reductionism, as opposed to systemic (holism), is a philosophical concept that refers both to the way of thinking solutions as well as to their modeling methodology. Reductionism advocates reducing system complexity or phenomenon to their basic elements which would then be easier to understand and study [18]. This reductionist approach, despite its high efficacy in several areas, shows, however, its limits within certain contexts. In fact, for explaining certain phenomena or solving certain problems, the approach consisting of reducing or abstracting the reality to a linearization of simple relationships of causes and effects between a complex system underlying fundamental components, appears as a highly limiting and simplifying approach.

With regard to schema matching, it seems clear, as Figure 2 illustrates it, that all current approaches follow the reductionist thinking. They abstract the matching process to a linear function with a set of inputs and outputs. This function is decomposed into a series of modules, each of which is responsible for the running of a stage of the process (e.g., selection and matching execution).

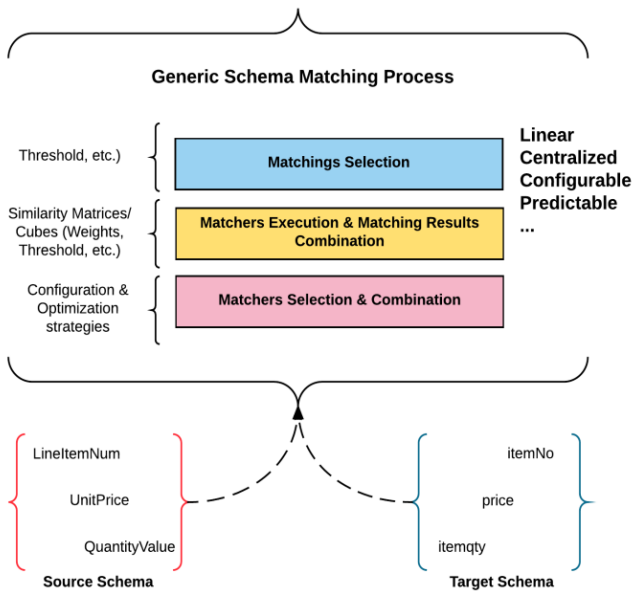

Figure 2. Generic Schema Matching process (linear process with an analytical-based resolution)

Some fundamental and intrinsic characteristics, common to all current Schema Matching systems, may partially explain their inability to overcome the limitation of the complexity and other challenges, such as uncertainty. Those characteristics are declined as following: these systems are (i) complicated and not complex, (ii) linear (analytical, deterministic and predictable) and not non-linear, (iii) centralized rather than decentralized (parallelism and emerging solutions), (iv) and finally, configurable and not adaptable (self-configuration, self-optimization).

The need to explore new approaches to make systemic and holistic responses to the problems of matching leads us to raise the question: how can we have a matching solution that could give us high-quality matching results, for different matching scenarios and this with a minimal optimization effort from the end-user?

Our premise is that a good part of the answer may come from the theory of CAS where modeling the complexity of adaptation and evolution of the systems is at the heart of this theory. Having a schema matching approach that can face and overcome the challenges facing the existing schema matching tools requires, in our view, a paradigm shift, placing the notions of adaptation, evolution, and self-organization at its center. We strongly believe that the theory of CAS, which is exploited to explain some biological, social, and economic phenomena, can be the basis of a programming paradigm for ASM tools.

### III. SCHEMA MATCHING AS A SYSTEMIC APPROACH

As part of our research we investigated the use of the theory of CAS (systemic thinking), to try to find an innovative response to challenges (i.e. complexity, uncertainty) that the conventional approaches for schema matching are still facing.

We think that the CAS could bring us the adaptation capability to the realm of schema matching tools (self-configuration and self-optimization), which should relieve the user from the complexity and effort resulting from configuring and optimizing the automatic schema matching systems.

Our conceptual model for schema matching, based on the theory of complexity, sees the schema matching process as a complex adaptive system.

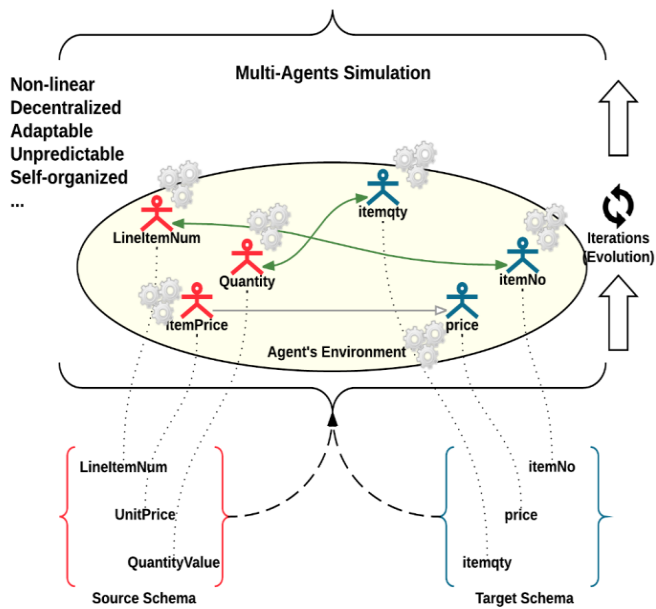

Figure 3. Schema Matching as Multi-Agents Simulation (non-linear process with emergence-based resolution)

As illustrated in Figure 3, in this model, each schema element of the schemas to match (source or target schema) is modeled as an autonomous agent, belonging to a population (source or target schema population). Each agent behaviors and interaction, at the micro level, with the other agents in the opposite population and with its environment, brings out at the macro level, a self-organized system that represents the global solution to matching problem (i.e., relationships between schemas elements). In other words, the resolution of the matching problem goes through individual effort deployed by each agent, locally, throughout the simulation to find the best match in the opposite population.

We think that many intrinsic properties of our model, derived from the ABMS modeling approach, can contribute efficiently to the increase of the matching quality and thus the decrease of the matching uncertainty. These properties are:
- Emergence: the emergence of the macro solution (schema matching) comes from local behaviors, rules and interactions between agents (micro solutions).
- Self-organization: the cooperation of source and target schema elements (represented as agents) helps to reach a consensus about their best matching.
- Stochasticity (randomness): the randomness within the model, gives the ability to perform statistical analysis on the outcome of multiple simulations (meta-simulation) for the same matching scenario.

Briefly, our idea is to model the Schema Matching process as interactions, within a self-organized environment, between agents called "Schema Attribute Agent". In the rest of the paper, we are going to refer to the "Schema Element Agent" simply as agent. Each schema element is modeled as an agent belonging to one of two populations: source or target schema group. Furthermore, the schema matching process is modeled as the interaction between the two populations of agents.

In our model, the internal architecture of the agents is Rule-based (reflexive agent). The agents have as a main goal to find the best matching agent within the other group of agents. The foundation of the rules governing the agent's behaviors is stochasticity (randomness). In fact, a certain degree of randomness is present in each step executed by each agent during the simulation.

The main random elements influencing the simulation are as follows:
- Similarity Calculation based on similarity measures selected randomly from a similarity measures list.
- Similarity Scores aggregation based on aggregation functions selected randomly from an aggregation function list (MAX, AVERAGE, WEIGHTED).
- Similarity score validation based on generated random threshold value (within interval)

As opposed to deterministic solutions for schema matching (all the existing matching solutions), the nondeterministic and stochastic nature of our agent-based simulation increase the confidence in the quality of the matching results. Even though the agent's behaviors are based on randomness (e.g., during the similarity calculation), our model can often produce the right matchings at the end of each simulation run.

Figure 4 illustrates the internal states of each agent. It allows representing the transitions between the internal states, during the perception-decision-action cycle of the agent. In the context of our operational model, the agent during the perception phase, perceives its environment by interrogating it, by performing similarity calculations (which can be considered as an act of recognition) or by capturing certain events. The result of this phase will be a set of percepts, allowing the agent to identify the agents of the other group, available for matching. The capture of events, coming from the environment, is another action of perception: for instance, the event that is triggered when the agent is chosen by another one as a matching candidate. During the decision phase, the agent from the results of the perception phase, reasons, deliberates and decides on the action to be selected. The decisions, involving the choice of actions, are the following: (i) the decision concerning the convergence of similarities and the selection of a candidate matching, (ii) the decision concerning the reset of the beliefs concerning the candidate matching, and (iii) the decision on consensual matching. During the action phase, the agent executes the actions selected during the previous phase. The current iteration of the simulation ends with this phase.

The behavior of the agent is driven by the goal of finding a consensual match. The consensus-selection approach is a naive approach, consisting of waiting for a consensus that must coincide for both agents (which may imply a longer duration for the simulation).

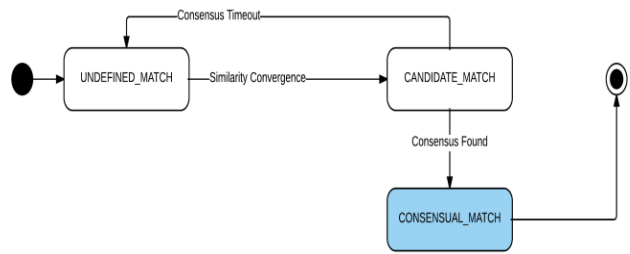

Figure 4. Agent's internal states

The main key-features of our conceptual model are summarized as follow:
- Stochastic Linguistic Matching: similarity calculation based on similarity measures selected randomly from a similarity measures list. Similarity Scores aggregation based on aggregation functions selected randomly from an aggregation function list (MAX, AVERAGE, WEIGHTED). Similarity score validation based on generated random threshold value (within interval).
- Consensual Matching Selection: to form a valid pairing/correspondence, the two agents (form opposite

populations: source and target schemas) should refer to each other as candidate match (in the same time).
- Meta-Simulations and Statistical Analysis: performing statistical analysis on multiple simulation runs data is a good way to improve the confidence in the matching result obtained from our model.

We believe that the conceptualization and the modeling of schema matching as multi-agent simulation will allow the design of a system exhibiting suitable characteristics:

(i) An easy to understand system, composed of simple reflexive "agents" interacting according to simple rules.

(ii) An effective and efficient system, autonomously changing over time, adapting, and self-organizing. A system allowing the emergence of a solution for any given matching scenario.

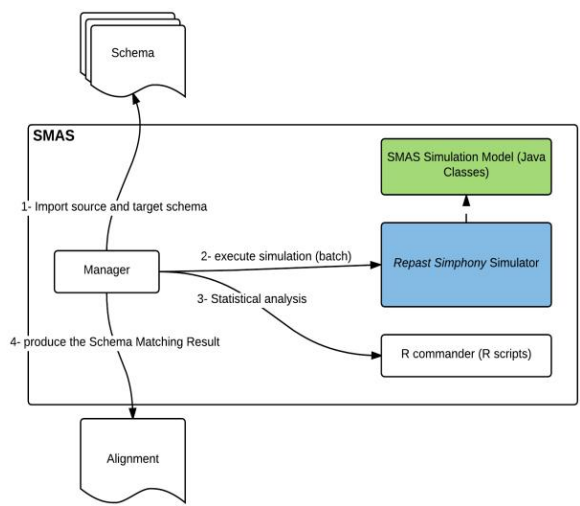

Figure 5. High-level Architecture for Reflex-SMAS

As depicted in Figure 5, our Reflex-SMAS prototype core was implemented in Java using the open source ABMS framework Repast Simphony (2.1) [19], and the open source framework for Text Similarity DKPro Similarity (2.1.0) [20]. The open source R language (R 3.1.0) [21] was used for statistical data analysis.

In the next section, we are going to describe the empirical evaluation of the prototype Reflex-SMAS.

## IV. EMPIRICAL EVALUATION

The validation of agent-based simulation models is a topic that is becoming increasingly important in the literature on the field of ABMS. Three types of validation could be identified [22]: (i) Empirical Validation, (ii) Predictive Validation, and (iii) Structural Validation.

As we will see in detail, the empirical validation is the type of validation that we have adopted for the evaluation of our Agent-based Simulation Model for Schema Matching (i.e. prototype Reflex-SMAS).

First, we will start with the description of the methodology used as our validation approach, and then we continue by providing a summarized view of our validation results.

### A. Evaluation Objectives and Strategy

We are seeking, through this empirical evaluation, to validate the following aspects of our prototype Reflex-SMAS:
- That our solution is, indeed, an effective and efficient automatic schema matching system, capable of autonomously changing behaviors and evolving over time, to adapt, and to self-organize and thus make the solution for any matching scenario to emerge.
- That our solution is easy to understand, and therefore, could display a high degree of maintainability (e.g., adding new matchers).

The proof strategy consists on conducting experiments and then collecting and analyzing data from these experiments. Thus, the validation approach that we have adopted is considered as a hybrid validation approach combining two validation approaches coming from two different fields, namely Schema Matching and ABMS. On one hand, from the field of Schema Matching, we are leveraging a popular evaluation method consisting of the comparison of results with those expected by the user [23]. On the other hand, from the field of ABMS, we are using the Empirical Validation [22] which is mainly based on the comparison among the results obtained from the model and what we can observe in the real system.

Thus, the strategy adopted for the validation of our prototype (implementing our multi-agent simulation model for schema matching) consists of:
- Defining different synthetic matching scenarios (three matching scenarios namely "Person", "Order" and "Travel") with different sizes and different level of lexical heterogeneity, so we can evaluate the prototype matching performance in different situations (adaptation).
- Conducting experiments, compiling results and evaluating the matching performance by comparing, for those three matching scenarios, the matching results (matchings) obtained from our prototype Reflex-SMAS with the results expected by the user.

In the first matching scenario "Person", we need to match two schemas with small size (i.e., six elements) showing a medium lexical heterogeneity level. The second matching scenario "Order" is composed of schemas with medium size with a high lexical heterogeneity level. The schemas in the last matching scenario "Travel" have a relatively big size with a low lexical heterogeneity level.

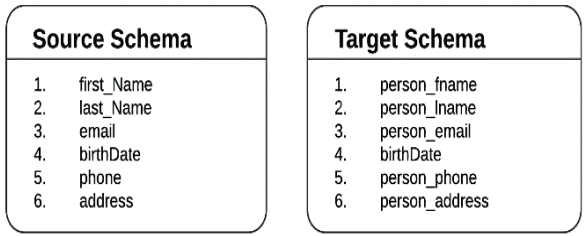

Figure 6. Matching Scenario "Person"

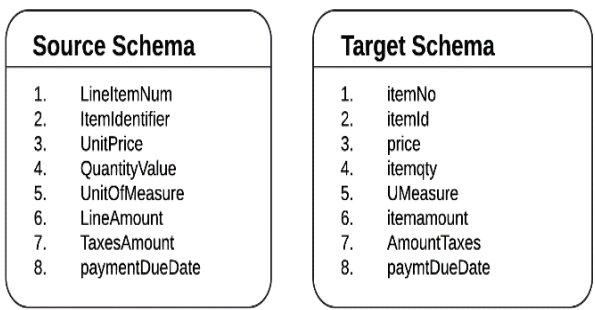

Figure 7. Matching Scenario "Order"

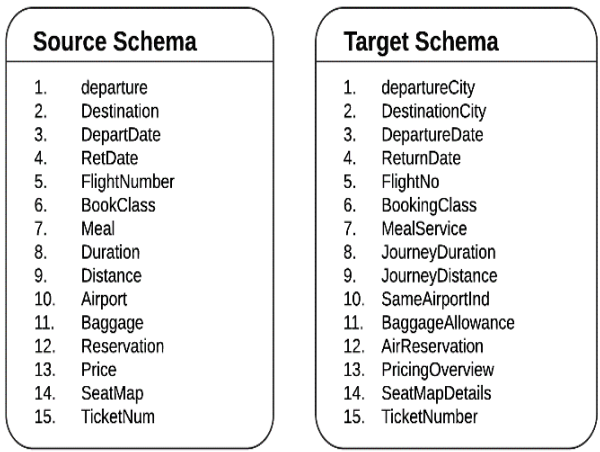

Figure 8. Matching Scenario "Travel"

In order to assess the relevance and level of difficulty that can represent those synthetic matching scenarios (i.e., "Person", "Order" and "Travel"), we decided to evaluate them, first, using the well-known matching tool COMA [24]–[26]. Since, the COMA tool was not able to resolve all the all expected matches for those scenarios, we can say that the proposed synthetic matching scenarios, should be enough challenging scenarios for our validation (from their level of heterogeneity perspective).

Regarding the experiments execution and results compilation, we have decided to run series of three meta-simulations for each scenario (each meta-simulation includes 10 simulations).

The final matching result is based on a statistical analysis of each meta-simulation outcome. In other word, the matching result relies on the calculation of the frequency of occurrence of a found match on the ten simulations composing the meta-simulation. Furthermore, executing for each scenario the meta-simulations three times is a choice that we made to help with the assessment of the experiment repeatability.

B. *Experiment Results*

This section summarizes the results obtained because of experiments conducted to evaluate the tool Reflex-SMAS.
After executing the set of three meta-simulations for each matching scenario, we have compiled the results for the performance for each meta-simulation for all scenarios. As indicated in Table I, our tool was able to correctly find all the expected correspondence by the user (a 100% success rate) after each meta-simulation, and for each scenario.

TABLE I. REFLEX-SMAS EXPERIMENT COMBINED RESULTS

| Scenario | M.S. | M. to F. | C.M.F. | % C.M.F. |
|---|---|---|---|---|
| Person | 1 | 6 | 6 | 100% |
| Person | 2 | 6 | 6 | 100% |
| Person | 3 | 6 | 6 | 100% |
| Order | 1 | 8 | 8 | 100% |
| Order | 2 | 8 | 8 | 100% |
| Order | 3 | 8 | 8 | 100% |
| Travel | 1 | 15 | 15 | 100% |
| Travel | 2 | 15 | 15 | 100% |
| Travel | 3 | 15 | 15 | 100% |

M.S: Meta Simulation
M. to F: Matchings to Find
C.M.F: Correct Matchings Found
% C.M.F: % Correct Matchings Found

Now, if we compare the results of our Reflex-SMAS prototype with COMA tool results, we can clearly notice that our tool outperformed the COMA tool in all the syntactic matching scenarios. Table II shows the compared result for Reflex-SMAS vs. COMA.

TABLE II. REFLEX-SMAS VS. COMA EXPERIMENT COMBINED RESULTS

| Scenario | M. to F. | Reflex-SMAS | | COMA | |
|---|---|---|---|---|---|
| | | C.M.F. | % C.M.F. | C.M.F. | % C.M.F. |
| Person | 6 | 6 | 100% | 5 | 83% |
| Order | 8 | 8 | 100% | 6 | 75% |
| Travel | 15 | 15 | 100% | 13 | 87% |

Figure 9 shows a comparison of the performance obtained for scenarios "Person", "Order" and "Travel" with our prototype compared to those obtained with the COMA tool.

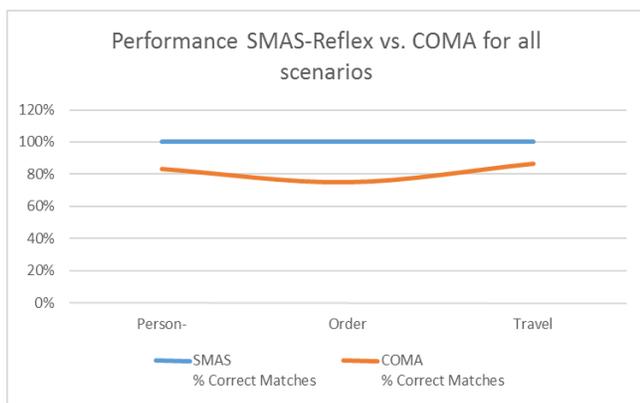

Figure 9. Comparative result between Reflex-SMAS and COMA

To challenge the "perfect" results obtained with our tool Reflex-SMAS for the synthetic matching scenarios, we were curious to know to what extent the performance obtained at the meta-simulations, may be impacted by a reduction in the number of individual simulations composing a meta-simulation. Therefore, we decided to conduct further experimentation, reducing, this time, the number of individual simulations of a meta-simulation from ten simulations to only three simulations.

As we can notice in Figure 10, the performance obtained in the experiment with the meta-simulations composed of three individual simulations instead of ten, has dropped for the scenarios "Order" and "Travel". It means that our matching tool Reflex-SMAS was not able to find all the expected matchings during some of the meta-simulations for those two scenarios (due to the high level of heterogeneity of the scenario "Order" and the big size of the scenario "Travel"). Unquestionably, we can conclude that the number of individual simulations, composing the meta-simulation is an important factor to ensure good matching performance (better quantification of the uncertainty regarding the outcome of the matching process) especially when it comes to scenarios involving large schemas and/or having a high level of heterogeneity.

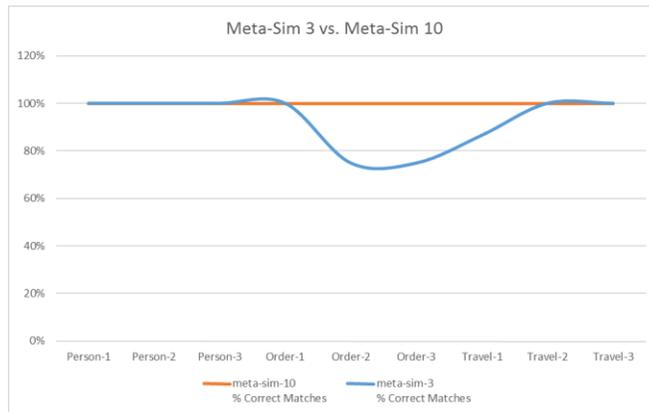

Figure 10. Meta-simulation with 3 vs. 10 individual simulations

V. CONCLUSION AND FUTURE WORKS

Our prototype (Reflex-SMAS) empirical evaluation showed us clearly its capability of providing a high-quality result for different schema matching scenarios without any optimization or tuning from the end-user. The experiments results are very satisfactory. Thus, we can conclude that approaching the schema matching as a CAS and modeling it as ABMS is a viable and very promising approach that could greatly help to overcome the problems of uncertainty and complexity in the field of schema matching.

Our approach represents a significant paradigm-shift, in the field of ASM. In fact, to the best of our knowledge, never the ASM problem has been addressed by adopting systemic thinking (holistic approach) or has been considered as a CAS and modeled using ABMS modeling approach.

As future work, we are planning to enhance the conceptual model of our prototype to tackle challenges, such as complex schema ($n{:}m$ cardinalities) by exploiting other Similarity Measures, such as Structural Similarities (schemas structures). On the other hand, in order to open up new perspectives and to overcome the limits of purely reactive behavior, we are thinking on a "conceptual" evolution of the internal architecture of our agent, evolving it from a reactive agent to an agent of rational type. This evolution consists in the implementation of a decision-making model under uncertainty, at the level of the decision-making phase of the agent, giving it the ability to reason and to choose between conflicting actions. The rational agent we are aiming for, should have a memory, a partial representation of its environment and other agents (its perception), and a capacity for reasoning, allowing it to make a rational choice (to choose the action with the greatest utility) that can guarantee it to maximize its satisfaction (measure of performance). The result of this conceptual evolution could give rise to a new version of our prototype.